\documentclass[letterpaper, 10 pt, conference]{ieeeconf}  

\IEEEoverridecommandlockouts                              
\usepackage{float} 
\usepackage{mathtools}
\usepackage{amssymb}
\usepackage{bbm}
\usepackage{graphicx}
\usepackage{multicol}
\usepackage{multirow}
\usepackage[bookmarks=true]{hyperref}
\usepackage{caption}
\usepackage{subcaption}
\usepackage{algorithmic}
\usepackage{algorithm}
\let\labelindent\relax
\usepackage{enumitem,kantlipsum}
\usepackage{tablefootnote}
\usepackage{booktabs}
\usepackage{dsfont}
\usepackage{gensymb}

\title{\LARGE \bf
Pixel-Attentive Policy Gradient for Multi-Fingered Grasping in Cluttered Scenes}

\author{Bohan Wu, Iretiayo Akinola and Peter K. Allen
\thanks{This work was supported in part by a Google Research grant and National Science Foundation grants CMMI-1734557 and IIS-1527747.
Authors are with Columbia University Robotics Group, 
        Columbia University, New York, NY 10027, USA
        {\tt\small  bw2505@columbia.edu, iakinola@cs.columbia.edu, allen@cs.columbia.edu}}
}

\begin{document}

\maketitle
\begin{abstract}
Recent advances in on-policy reinforcement learning (RL) methods enabled learning agents in virtual environments to master complex tasks with high-dimensional and continuous observation and action spaces. However, leveraging this family of algorithms in multi-fingered robotic grasping remains a challenge due to large sim-to-real fidelity gaps and the high sample complexity of on-policy RL algorithms. This work aims to bridge these gaps by first reinforcement-learning a multi-fingered robotic grasping policy in simulation that operates in the pixel space of the input: a single depth image. Using a mapping from pixel space to Cartesian space according to the depth map, this method transfers to the real world with high fidelity and introduces a novel attention mechanism that substantially improves grasp success rate in cluttered environments. Finally, the direct-generative nature of this method allows learning of multi-fingered grasps that have flexible end-effector positions, orientations and rotations, as well as all degrees of freedom of the hand.
\end{abstract}

\section{INTRODUCTION}
Multi-fingered grasping remains an active robotics research area because of the wide application of this skill in different domains, ranging from using robot arms for warehouse handling to using humanoid robots for home assistant robotic applications. There are different approaches to robotic grasping; largely categorized into \textit{classical grasp planning} approaches that optimize closed grasp quality metrics and \textit{learning-based} methods that learn from examples or experience. The data-driven methods have become more popular in recent years as they leverage many recent advancements in the deep learning community. 

We observe that the majority of the learning based methods employ low-DOF robot hands (e.g. parallel-jaw) and also often limit the range of the grasp approach direction (e.g. top-down grasps). While these two DOF restrictions reduce the dimensionality of the problem, they exclude many solutions that could be used for applications like semantic grasping or grasping for manipulation. For example, top-down grasping of a bottle/cup could hamper a pouring manipulation task. In addition, since a full 6-DOF grasping system subsumes the more popular 4-DOF methods, the learned system can be left to decide if 4-DOF system will be sufficient based on the given grasping situation. The choices of the learned algorithm can be analyzed to see which scenarios resulted in reduced-DOF grasps versus other grasp poses. This leaves the debate of whether a reduced-DOF grasping system is sufficient entirely to the learned algorithm.

\begin{figure}[h]
\vspace{-4mm}
    \centering
    \includegraphics[width=0.9\columnwidth]{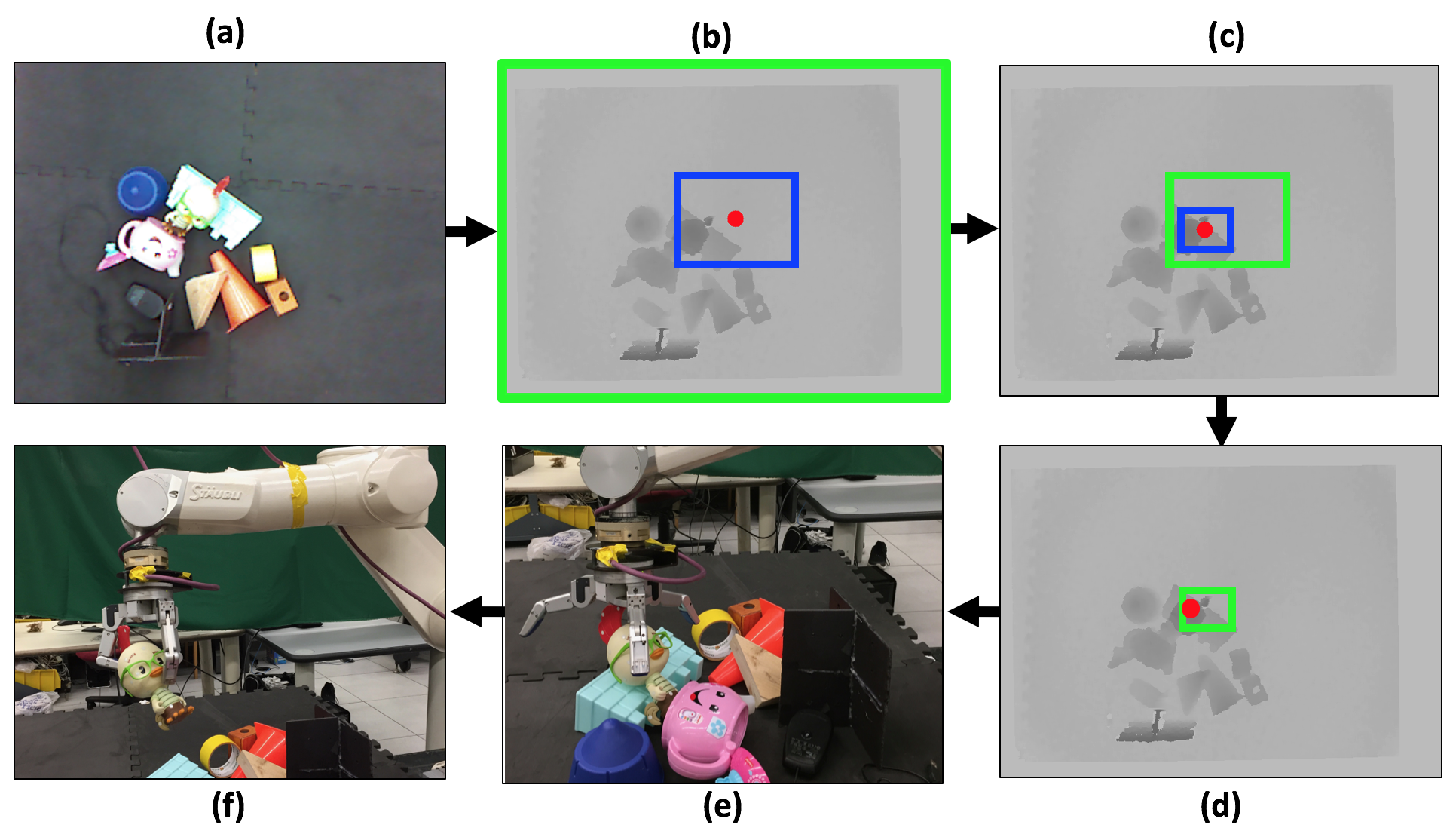}
    \caption{\small \textbf{Pixel-Attentive Policy Gradient Multi-Fingered Grasping.} Given a scene of cluttered objects \textbf{(a)}, our method takes in a single depth image and gradually zooms into a local region of the image to generate a good grasp. \textbf{(b)}, \textbf{(c)} and \textbf{(d)} show the zooming process, in which the green bounding box represents the portion of the depth image the robot observes in the current timestep, and the blue bounding box represents the portion of the depth image the robot wants to observe in the next timestep. In the end, a full-DOF grasp is learned based on the final zoomed image \textbf{(d)} as shown in \textbf{(e)} and with the final pick-up shown in \textbf{(f)}.}
    \label{zoom}
\vspace{-2mm}
\end{figure}
In this work, we address a fundamental paradox in learning-based grasping systems: attempting to increase the robot's DOFs to fully capture the robot's action capacity versus the competing objective of keeping the sample complexity (i.e. the amount of training data needed to learn a good grasp) manageable. While including more DOFs in grasping robots, such as allowing non-top-down grasping and using multi-fingered hands, can increase their potential to perform better grasps, it also increases the complexity of the problem. The increased complexity affects the stability of many learning algorithms during training especially for continuous action spaces and their effectiveness/robustness when transferred to real world settings.
Currently, policy gradient methods solve this paradox well \textit{usually in simulation}. By combining advanced policy optimization procedures with neural-network functional approximators, this family of algorithms can solve complex tasks in simulation with high-dimensional observation and action spaces \cite{schulman2017proximal}. While these methods can capture the potential of higher action spaces, the on-policy nature of policy gradient methods requires a level of sample complexity that is almost infeasible in physical environments without large-scale parallelized robotic systems \cite{kalashnikov2018qt}\cite{levine2017learning}. In addition, the brittle nature and complex manifold of robotic grasping where a slight perturbation of a good solution can result in a very bad grasp \cite{rosales2011global} means that optimizing in higher dimensions is more complex.

This work tackles the inherent increase in sample complexity for grasping with multi-fingered hands. We introduce a method that learns the finger joint configurations for high DOF hands (i.e. $\#DOFs \geq 2$) and full grasp pose (3D position and 3D orientation) that will return successful grasps. Our novel architecture takes as input a single depth image of a cluttered or non-cluttered scene and reinforcement-learns a model that produces a flexible grasp pose and all finger joint angles. Each component of this grasp is proposed per-pixel and then converted into the appropriate space. For example, the grasp position output by our grasping policy is in the depth image's pixel space and later converted to Cartesian space using the pixel-to-Cartesian mapping (point cloud) inferred from the depth image.

\begin{figure}[t]
\vspace{3mm}
\centering
    \begin{subfigure}[t]{0.1675\paperwidth}\includegraphics[width=0.94\linewidth]{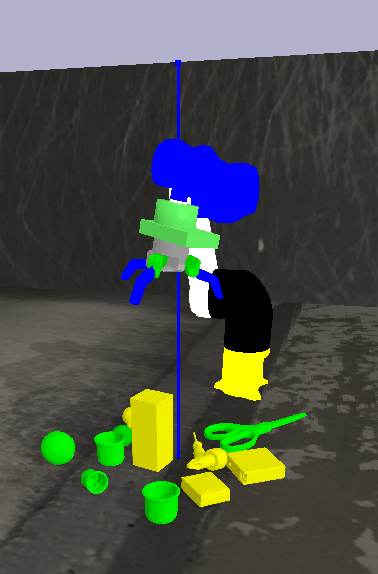}\caption{PyBullet (Train)}\label{fig:pb}\end{subfigure}
    \begin{subfigure}[t]{0.1913\paperwidth}\includegraphics[width=0.94\linewidth]{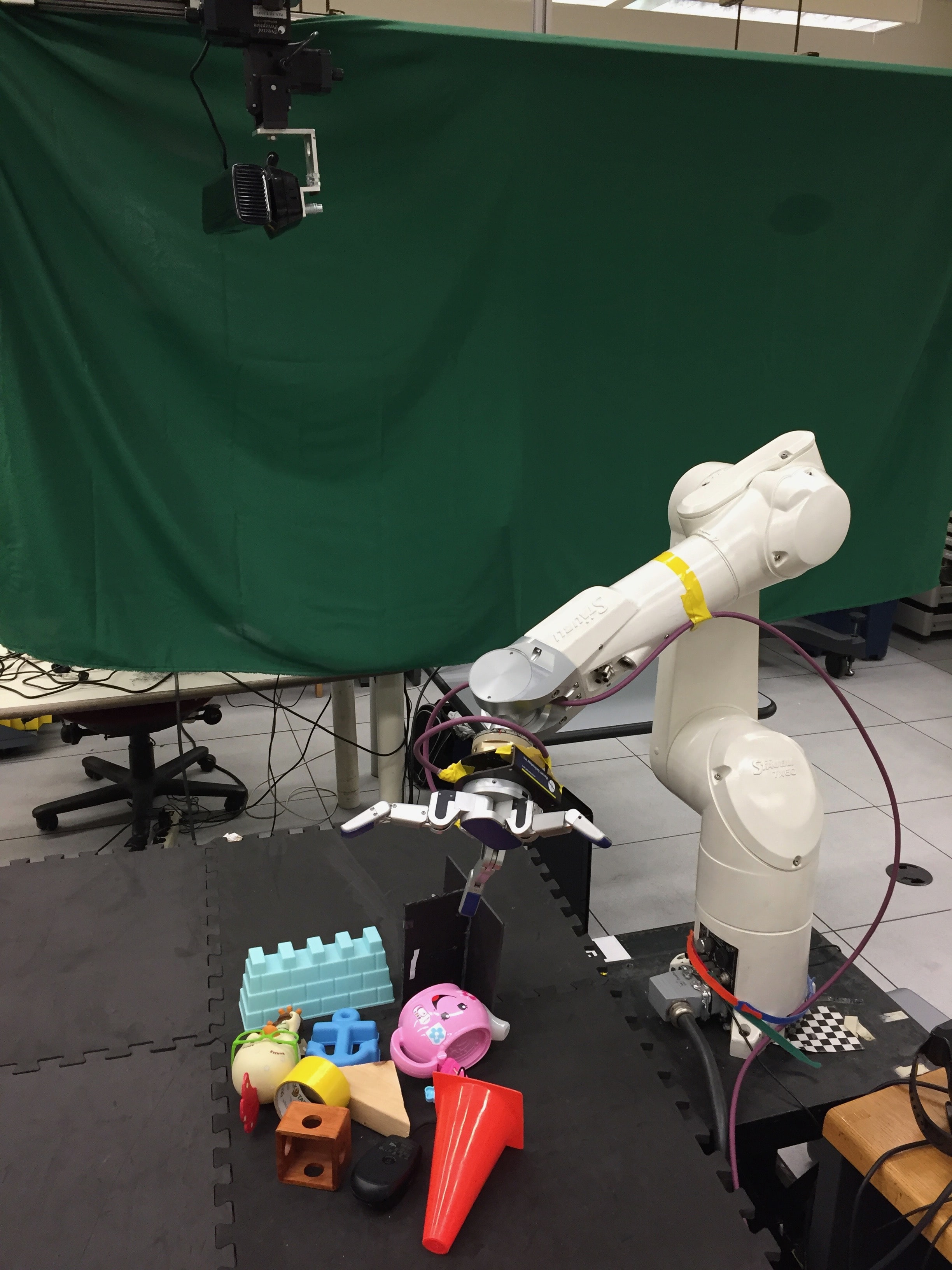}\caption{Robot (Test)}\label{fig:robot}\end{subfigure}
    \caption{\small Hardware and Simulation Setup. \textbf{(a)} A sample grasping scene in PyBullet simulation used to train our grasping policy. \textbf{(b)} Example real world grasping scene for testing the trained algorithm.}
    \label{fig:hardware}
\vspace{-5mm}
\end{figure}

Learning an RL policy operating directly in the image pixel space not only optimizes sample efficiency during sim-to-real transfer but also enables our novel attention mechanism that learns to focus on sub-regions of the depth image in order to grasp better in cluttered environments (hence the term ``pixel-attentive'').  Shown in Figure~\ref{zoom}, the proposed mechanism optionally crops the depth image sequentially to gradually zoom into a local region of the scene with a higher chance of generating good grasps. This mechanism also learns to stop cropping the image after enough zooming and generate non-top-down grasps with variable end-effector positions, orientations and all finger joint angles. In summary, our contributions are:
\begin{enumerate}
    \item An RL algorithm that can solve cluttered grasping scenes with full DOFs: flexible end-effector position, orientation and finger joint angles using multi-fingered robotic hands
    \item A novel attention feature that allows the robot to understand cluttered scenes and focus on favorable grasp candidates using a zoom-in mechanism
    \item Simulation-based learning that uses depth and geometry alone (i.e. no texture) to allow accurate domain transfer to real scenes
    \item Multiple experiments both in simulation and real-world scenes that produce very high levels of grasping success
\end{enumerate}

\section{RELATED WORK}
\subsection{Learning Grasping under Sample Complexity Challenges}
Advancements in deep learning have given rise to the rapid development of learning-based techniques for robotic grasping. \cite{bohg2014data} gives a review of learning-based grasping methods. Some of these techniques use a supervised learning approach where a model is trained on grasping examples \cite{lenz2015deep}\cite{morrison2018closing}. On the other hand, there are also RL-based techniques that train a grasping policy based on trial and error-- learning to perform actions resulting in grasp success. A major issue common to both supervised and RL methods is the challenge of sample complexity. A large amount of data-- ranging from thousands to millions-- is required for a majority of the data-hungry techniques. However, real world robotics data are very expensive to collect, with labelled data even more expensive. 
Increase in sample complexity can result from curse of input or action dimensionality, dealing in continuous spaces instead of discrete spaces, increase in neural network capacity, and change in learning paradigm (RL vs. supervised). For example, learning a 6-DOF grasp pose for a multi-fingered hand will likely require much more data and is more susceptible to learning stability issues than learning a 4-DOF grasp pose for a parallel-jaw hand.

To tackle the challenges associated with sample complexity in grasping, recent works attempted a variety of algorithms, frameworks and procedures. The first branch of attempts avoids on-policy RL methods and uses alternative algorithms with lower sample complexity, such as supervised convolutional neural networks (CNN) \cite{morrison2018closing}\cite{varley2015generating}, value-function based deep RL such as Deep Q-learning \cite{kalashnikov2018qt}\cite{quillen2018deep}\cite{zeng2018learning}, RL with a mixture of on-policy and off-policy data \cite{kalashnikov2018qt}, and imitation learning \cite{hsiao2006imitation}. The second branch uses various procedures to limit the search space of the learning algorithm. For example, one can leverage the image-to-coordinate mapping based on the point cloud computed from the camera's depth image so that the algorithm can only learn to choose a point in the point cloud from the image as opposed to the desired 3D position in the robot's coordinate system \cite{morrison2018closing}\cite{varley2015generating}, a philosophy that inspired our approach. Alternatively, one can restrict the robot's DOFs to top-down grasps only \cite{morrison2018closing}. The third branch learns grasping in simulation and proposes sample-efficient sim-to-real frameworks such as domain adaptation \cite{bousmalis2018using}, domain randomization \cite{tobin2017domain}, and randomized-to-canonical adaptation \cite{james2018sim} to transfer to the real world.

\subsection{Vision-Based Grasping}
Real-world grasping requires visual information about the environment and the graspable objects to generate grasps. This can be RGB \cite{kalashnikov2018qt}, RGB-D \cite{varley2015generating}\cite{zeng2018learning}, or depth-only data \cite{varley2017visual}. In this work, we only use a single depth image as input and avoid RGB information due to two considerations. First, texture information could be less useful than local geometry in the problem of grasping. Second, RGB images are arguably harder to simulate with high fidelity than depth maps and using them increases the sim-to-real gap. The debate on how best to bridge the domain gap between simulation and real world remains active and there is no consensus on which approach works best.

\textbf{\textit{Attention Mechanism}}: This refers to a class of neural network techniques that aims to determine which regions of an input image are important to the task at hand. By applying convolutional or recurrent layers to the input, a saliency map is generated; it has a per-pixel score proportional to the importance of each location in the image. While previous works have applied this mechanism to saliency prediction \cite{pan2016shallow}, we use attention to enable improved grasping in cluttered scenes. A previous work \cite{wang2018deep} used the attention mechanism to determine a cropping window for an input image that maximally preserves image content and aesthetic value. We apply this idea to predict which region of a cluttered scene to zoom-in on towards achieving improved robotic grasping success.
We set-up this attention-zooming mechanism in a fully reinforcement-learned manner.

\subsection{Multi-Fingered Grasping}
Multi-fingered grasping, commonly referred to as \textit{dexterous grasping} in the literature, has been tackled using classical grasping methods \cite{berenson2008grasp}. These methods use knowledge of the object geometry and contact force analysis to generate finger placement locations in a way that encloses the target object and is robust to disturbances, i.e. the object stays in hand under environmental disturbances \cite{okamura2000overview}. To achieve this, some methods sequentially solve for each finger location where the placement of a new finger depends on the placement of the previous placed ones \cite{hang2014hierarchical}. On the other hand, some methods reduce the dimensionality of the finger joints into a smaller set of grasp primitives so the grasp search/optimization is done in a smaller subspace \cite{berenson2008grasp}\cite{ciocarlie2009hand}\cite{saut2007dexterous}\cite{ciocarlie2007dexterous}. Deep learning for dexterous grasping is less popular. \cite{varley2015generating} developed a supervised learning method that proposes heatmaps for finger placement locations in pixel space which guide a subsequent grasp planning stage, which is essentially a hybrid based method. More recently, \cite{schmidt2018grasping} proposed a fully learned approach that predicts 6D grasp pose from depth image. Their method uses supervised learning and requires a dataset of good grasps to train on. In contrast, we take an RL approach that learns to grasp more successfully via trial and error.

In summary, our method takes in a depth image and uses a policy gradient method to predict full 6-DOF grasp pose and all finger joint angles. \cite{tobin2018domain} presented an auto-regressive approach that can be extended to learn full-DOF grasp pose. However, they only show top-down grasp results for parallel-jaw hands. Another closely related work \cite{viereck2017learning}, trained on simulated depth images, proposed a supervised learning method that predicts grasps based on input depth image obtained from a hand-mounted camera. Their method greedily moves the gripper towards the predicted grasp pose as new images are continuously captured. In contrast to their method, ours does not require moving the robot arm to take a closer shot of the scene; instead, we capture the depth image only once and use a learned attention mechanism to shift focus and zoom into the image to a level that will maximize grasp success. To the best of our knowledge, our work is the first to propose an RL grasping algorithm for full-DOF grasp pose, all finger joint angles and multi-fingered hand. 

\section{Preliminaries}
\subsection{RL Formulation for Multi-Fingered Grasping}
We assume the standard RL formulation: a grasping robotic \textit{agent} interacts with an \textit{environment} to maximize the expected reward~\cite{sutton1998reinforcement}.
The environment is a Partially Observable Markov Decision Process, since the agent cannot observe 1) RGB information or 2) the complete 3D geometry of any object or the entire scene. To foster good generalization and transfer of our algorithm, we model this environment as an MDP defined by $\langle \mathcal{S}, \mathcal{A}, \mathcal{R}, \mathcal{T}, \gamma \rangle$ with an observation space $\mathcal{S}$, an action space $\mathcal{A}$, a reward function $\mathcal{R}: \mathcal{S} \times \mathcal{A} \to \mathbb{R}$, a dynamics model $\mathcal{T}: \mathcal{S} \times \mathcal{A} \to \Pi(\mathcal{S})$, a discount factor $\gamma \in [0, 1)$, and an infinite horizon.
$\Pi(\cdot)$ defines a probability distribution over a set.
The agent acts according to stationary stochastic policy $\pi: \mathcal{S} \to \Pi(\mathcal{A})$, which specify action choice probabilities for each observation.
Each policy $\pi$ has a corresponding $Q_\pi: \mathcal{S}\times\mathcal{A} \to \mathbb{R}$ function that defines the expected discounted cumulative reward for taking an action $a$ from observation $s$ and following $\pi$ from that point onward.

\subsection{Hardware and Simulation Setup}
We use the Barrett Hand (BH-280) mounted on a Staubli-TX60 Arm in both real-world and PyBullet \cite{coumans2016pybullet} simulation (Figure~\ref{fig:pb}). In the real-world (Figure~\ref{fig:robot}), a Kinect Depth Camera is mounted statically on top of the grasping scene.

\section{Pixel-Attentive Multi-fingered Grasping}
\begin{figure*}[t]
    \centering
    \includegraphics[width=1.8\columnwidth]{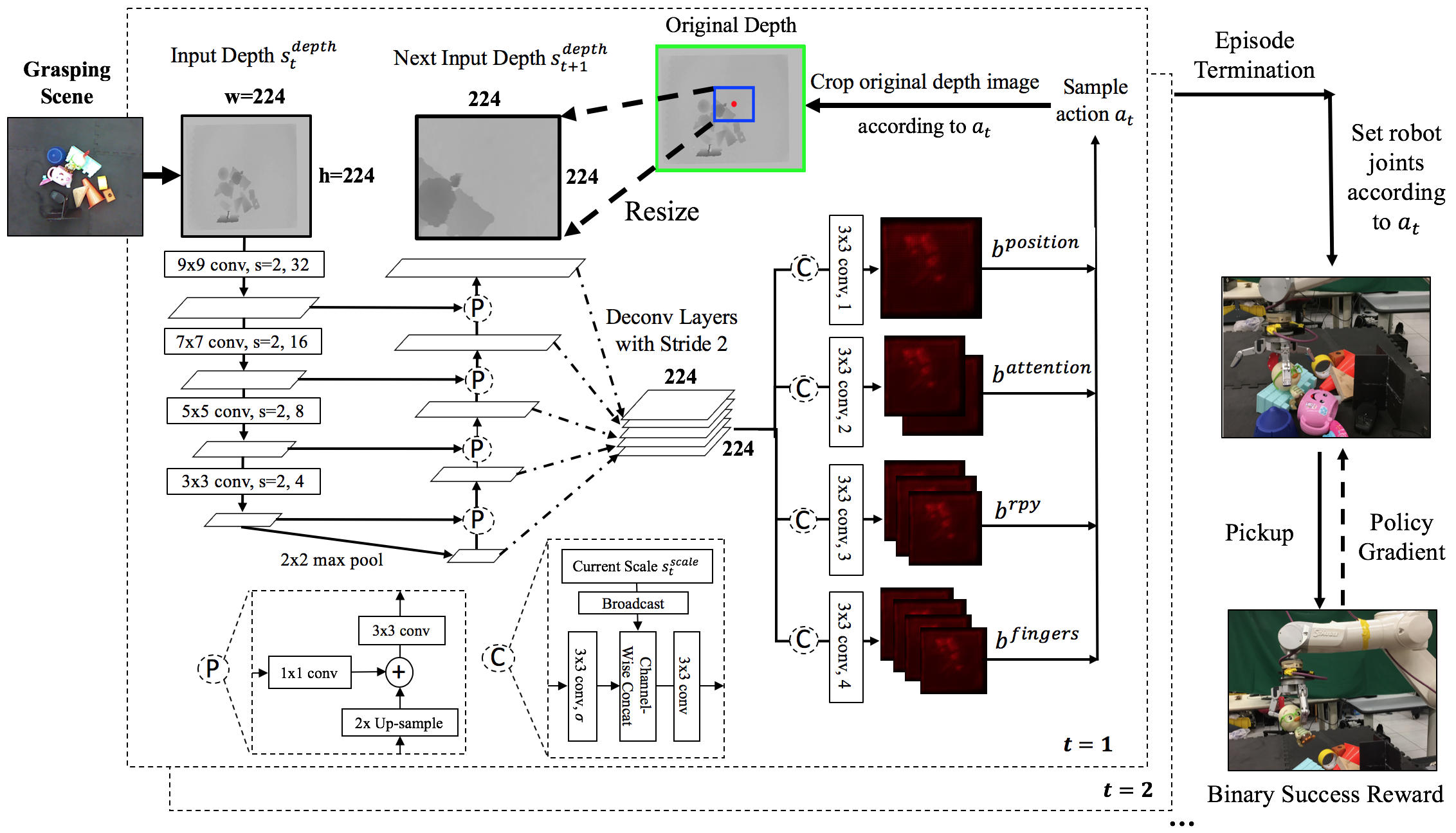}
    \caption{\small \textbf{Pixel-Attentive Multi-fingered Grasping architecture}. The $224 \times 224$ input depth map of a grasping scene $s_t^{depth}$ is accepted as input (top left) into a feature-pyramid four-branch CNN that outputs 10 activation maps. The ``P'' blocks indicate feature pyramid blocks, giving scale invariance capability to the CNN. The ``C'' blocks indicate how the current zoom-level $s_t^{scale}$ is introduced into each branch. All convolutional layers have ReLU activations and strides of 1 unless otherwise specified. For example, ``9x9 conv, s=2, 32'' refers to 9x9 kernel, stride of 2, 32 channels and ReLU activation. The number of deconvolutional layers ranges from 1 to 5 to upsample various intermediate feature maps back to $224 \times 224$. Each red-black map proposes pixel-wise grasp configurations and their probabilities. An action encoding the end-effector position, orientation and all finger joint angles is sampled from these maps, as well as a learned binary flag to decide whether to zoom further into the depth map at a certain scale or stop zooming and start grasping. If the decision is to zoom, the original depth map is cropped (blue bounding box on ``Original Depth'' image) according to the sampled action, resized and fed back into the CNN for the next timestep $t+1$, forming ``attention''. As the episode terminates, a binary success reward is acquired from the environment and policy gradient gets back-propagated into the entire network.}
    \label{algo}
\vspace{-5mm}
\end{figure*}

Our approach models the task of multi-fingered grasping as an infinite-horizon MDP. During each episode, the robot makes a single grasp attempt on the scene. During each timestep $t$ of the episode, the robot either 1) zooms into a local region of the scene via a reinforcement-learned attention mechanism or 2) terminates the episode and attempts a grasp based on the current zoom-level.

To begin, during the first timestep $t=1$, a single depth image of the grasping scene is first captured by a depth camera and resized to $224 \times 224$: $s_t^{depth} \in \mathbb{R}^{224 \times 224}$. This depth image, along with a scalar ratio indicating the current image's zoom-level $s_t^{scale} \in \mathbb{R}$, serves as the robot's observation: $s_t = \{s_t^{depth}, s_t^{scale}\} \in \mathbb{R}^{224 \times 224 + 1}$. The scalar ratio $s_t^{scale}$ gives the robot an ability to gauge the actual size of the objects during zooming or grasping and is initially 1 since no zooming was previously performed.
Next, both the depth image and the current zoom-level are fed into a four-branch CNN $f$, which has a shared encoder-decoder backbone $g_{backbone}$ and four branches $\{b^{position}, b^{attention}, b^{rpy}, b^{fingers}\}$:
\begin{equation}
    \begin{split}
        f^{x}(s_t) &= b^{x}(g_{backbone}(s_t^{depth}), s_t^{scale}),\\ \forall x &\in \{position, attention, rpy, fingers\}
    \end{split}
\end{equation}
This four-branch CNN outputs 10 two-dimensional (2D) maps, which define a single action $a_t$ encoded by 10 scalars: 
\begin{equation}
\begin{split}
a_t = \{a^{position}_t&, a^{zoom}_t, a^{scale}_t, a^{roll}_t, a^{pitch}_t, a^{yaw}_t, \\&a^{spread}_t, a^{finger1}_t, a^{finger2}_t, a^{finger3}_t\}\\
\end{split}\raisetag{3\baselineskip}
\end{equation}\label{eq:2}
\begin{align*}
\text{where }f^{position}(s_t) &\rightarrow a^{position}_t,\\
f^{attention}(s_t) &\rightarrow \{a^{zoom}_t, a^{scale}_t\}, \\
f^{rpy}(s_t) &\rightarrow \{a^{roll}_t, a^{pitch}_t, a^{yaw}_t\},\\
\text{and }f^{fingers}(s_t) &\rightarrow \{a^{spread}_t, a^{finger1}_t, a^{finger2}_t, a^{finger3}_t\}
\end{align*}

Given this action $a_t$, the robot either zooms into a local region of the depth map (blue bounding box in Figure~\ref{algo}) or directly performs a fully-defined grasp. The ``attention'' actions $\{a_t^{zoom},a_t^{scale}\}$ allow the robot to pay attention to a local region of the scene to make better grasps, therefore we term this local region the robot's ``region of attention''.  Among the 10 action scalars:
\begin{enumerate}
    \item $a^{position}_t$ represents the robot's end-effector position during grasping and the center location of the robot's region of attention during zooming;
    \item $a^{zoom}_t$ and $a^{scale}_t$ represent the zoom vs. grasp decision flag and the scale of the zooming respectively;
    \item $a^{roll}_t$, $a^{pitch}_t$ and $a^{yaw}_t$ represent the roll, pitch, yaw of the end-effector respectively during grasping;
    \item $a^{spread}_t$, $a^{finger1}_t$, $a^{finger2}_t$, and $a^{finger3}_t$ represent the lateral spread, finger-1, finger-2, and finger-3 pre-grasp joint angles of the 4-DOF Barrett hand during grasping.
\end{enumerate}
Below we discuss each of them in detail. 
\subsection{The Position Map}
The Position Map $f^{position}$ encodes the robot's end-effector position during grasping and the center location of the robot's region of attention (red dot in Figure~\ref{algo}) during zooming. Instead of encoding this position/location in Cartesian coordinates $\{x, y, z\}$, which will result in a very large and continuous action space to learn from, we observe that effective grasp positions can be associated with a point in the scene's point cloud, which is a discrete and smaller action space. Therefore, we encode this position/location using a single-channel 2D map of logits for a spatial-softmax distribution \cite{levine2016end}: $f^{position}: S \to \mathbb{R}^{224 \times 224}$, from which a pixel location $a_t^{position}$ can be sampled:
\begin{equation}
\begin{split}
    a_t^{position} &\sim \pi (a_t^{position} \mid s_t) \\
    &= \operatorname{spatial-softmax}(logits=f^{position}(s_t)) \\
    &\in [1, 224 \times 224]
\end{split}
\end{equation}
Given this pixel location $a_t^{position}$: 
\begin{enumerate}
    \item if the robot decides to \textit{zoom}, a bounding box centered around $a^{position}_t$ with a scale determined by $a_t^{scale}$ is cropped from the original depth map and resized back to $224 \times 224$. The resulting image becomes $s_{t+1}^{depth}$: the input depth map for the next timestep $t+1$;
    \item if the robot decides to \textit{grasp}, $a_t^{position}$ represents a unique point in the point cloud; the depth value at this pixel location $a_t^{position}$ is converted to an $(x, y, z)$ Cartesian location that the end-effector will be located before closing its fingers and trying to grasp.
\end{enumerate}
Because this pixel location enables the robot to zoom into the robot's region of attention and place the end-effector on a local point, we term this pixel location $a_t^{position}$ the robot's ``point of attention''.
Whether the robot decides to zoom in or grasp depends on the output of the Attention Maps, which we discuss in the next section.

\subsection{The Attention Maps}
The Attention Maps $f^{attention}$ make two decisions. First, they decide whether the robot should 1) zoom further into the depth map or 2) stop zooming further and start grasping. Second, they determine the level of zooming the robot should perform to acquire a better grasp down the road if the first decision is to zoom rather than grasp. These two decisions are important for grasping in cluttered scenes because while zooming into a cluttered scene can enable the robot to pay attention to a less visually-cluttered environment, too much zooming can cause the robot to lose sight of nearby objects. 

In addition, these two decisions should be different for different points of attention $a_t^{position}$. For example, if the current point of attention corresponds to a 3D point located on top of an object, then grasping could be a better decision than zooming. On the contrary, if the current point of attention corresponds to a 3D point located on the table where the objects reside, then zooming could be a better decision than grasping. Similar reasoning applies to the zoom-level. Therefore, instead of encoding these two decisions as two one-size-fits-all scalars, we use a two-channel map where each pixel on the map represents how much the robot intends to zoom vs. grasp and the zoom scale for every possible point of attention: $f^{attention}: S \rightarrow \mathbb{R}^{224 \times 224 \times 2}$. The first value on each pixel is the $p$ parameter for a Bernoulli distribution, and the robot makes the zoom vs. grasp decision $a_t^{zoom}$ by sampling a binary digit from this distribution:
\begin{equation}
    \begin{split}
        a_t^{zoom} 
        &\sim \pi (a_t^{zoom} \mid s_t, a_t^{position}) \\&= \operatorname{Bern} (\operatorname{sigmoid}({f^{attention}(s_t)_{(a_t^{position}, 1)}}))\in \{0, 1\}
    \end{split}\raisetag{2\baselineskip}
\end{equation}

If $a_t^{zoom} = 1$, the robot zooms further into the depth map. If $a_t^{zoom} = 0$, the robot stops zooming and makes the grasp.
The second value on each pixel represents the sigmoid-activated mean of a Gaussian distribution from which the robot samples the zoom scale $a_t^{scale}$. This zoom scale is a scalar that represents the height/width of the desired region of attention as a fraction of the current image size ($224 \times 224$), while the height/width aspect ratio remains the same:
\begin{equation}
    \begin{split}
        a_t^{scale} 
        &\sim \pi (a_t^{scale} \mid s_t, a_t^{position}) \\&= \operatorname{sigmoid}({f^{attention}(s_t)_{(a_t^{position}, 2)}}) \in [0, 1]
    \end{split}
\end{equation}

\subsection{The RPY(Roll-Pitch-Yaw) Orientation Maps}
The RPY Orientation Maps $f^{rpy}$ determine the end-effector orientation of a grasp by specifying the roll, pitch and yaw rotations from the unit vector $[1, 0, 0]$. Similar to the case of the Attention Maps, the RPY values ought to be different for different points of attention $a_t^{position}$. For example, if the current point of attention corresponds to a 3D point located on the top of an object, then a good set of RPY values should correspond to a near-top-down grasp. On the contrary, if the current point of attention corresponds to a 3D point located on the front of the object, then a good set of RPY values should correspond to a more forward-facing grasp. Therefore, instead of representing the three RPY values using three one-size-fits-all scalars $\{\alpha, \beta, \gamma\}$ across all possible points of attention, we do so using a three-channel map where each pixel on the map determines the three RPY values for every possible point of attention: $f^{rpy}: S \to \mathbb{R}^{224 \times 224 \times 3}$.

To determine each of the three RPY components $rpy_i \in \{\text{roll}, \text{pitch}, \text{yaw}\}$ for each $a_t^{position}$, the robot samples from a Gaussian distribution, whose mean $\mu_{rpy_{i}}$ is determined by the per-channel value of the pixel at the corresponding $a_t^{position}$ and whose standard deviation $\sigma_{rpy_{i}}$ is determined by a learned scalar parameter across all possible $a_t^{position}$:
\begin{equation}
    \begin{split}\label{eq:ori}
    a_t^{rpy_{i}} 
    &\sim \pi (a_t^{rpy_{i}} \mid s_t, a_t^{position}) \\
    &= \mathcal{N}(\mu_{rpy_{i}}, \sigma_{rpy_{i}}) \times \pi\\
    &= \mathcal{N}(\operatorname{activation_i}({f^{rpy}(s_t)_{(a_t^{position}, i)}}), \sigma_{rpy_{i}}) \times \pi
\end{split}\raisetag{2\baselineskip}
\end{equation}

For each $rpy_{i}$ orientation component (roll, pitch, yaw), the activation functions are (tanh, sigmoid, tanh). This results in an effective range of ($[-\pi, \pi]$, $[0, \pi]$, $[-\pi, \pi]$) respectively. Note that the pitch angle range is $[0, \pi]$ as opposed to $[-\pi, \pi]$ because only pitch values within $[0, \pi]$ produce meaningful grasps with the end-effector facing downwards (but not necessarily top-down) as opposed to upwards.

\subsection{The Finger Joint Maps}
The four finger joint maps $f^{fingers}$ determine the pre-grasp finger joint positions of the Barrett hand \textit{before} closing all fingers with the same joint velocity. Each of the four maps represents the lateral spread, finger-1, finger-2, and finger-3 pre-grasp joint angles of the under-actuated Barrett hand: $f^{fingers}: S \to \mathbb{R}^{224 \times 224 \times 4}$. Note that this formulation naturally extends to hands with more DOFs.

To propose angle for each of the hand joints $joint_i \in \{\text{spread}, \text{finger1}, \text{finger2}, \text{finger3}\}$ given $a_t^{position}$, each joint angle $a_t^{joint_{i}}$ is sampled from a Gaussian distribution and then scaled by the scaling factor $scale_{joint_{i}}$. The mean of this Gaussian distribution $\mu_{joint_{i}}$ is determined by the sigmoid-activated value of the corresponding map at $a_{position}$ and the standard deviation $\sigma_{joint_{i}}$ is a learned scalar parameter across all possible $a_{position}$:
\begin{equation}\label{eq:finger_joint_a}
    \begin{split}
        a_t^{joint_{i}} 
        &\sim \pi (a_t^{joint_{i}} \mid s_t, a_t^{position}) \\
        &= scale_{joint_{i}} \times \mathcal{N}(\mu_{joint_{i}}, \sigma_{joint_{i}})\\
         &= scale_{joint_{i}} \\ &\;\;\;\;\times \mathcal{N}(\operatorname{sigmoid}({f^{fingers}(s_t)_{(a_t^{position}, i)}}), \sigma_{joint_{i}})
    \end{split}\raisetag{3\baselineskip}
\end{equation}
For the Barrett hand used, the $scale_{joint_{i}}$ is $\pi/2$ for the lateral spread joint and $0.61$ for each of the other 3 finger joints. This gives an effective range of [0, $\pi/2$] for the lateral spread and [0, 0.61] for the 3 finger joints.
We restrict the finger-1, 2, and 3 joint ranges to be a quarter of the maximum range $[0, 2.44]$ because outside of this range the hand is nearly closed. We restrict the lateral spread to $[0, \frac{\pi}{2}]$ because outside of this range no meaningful grasps can be generated (all fingers will be on the same side of the hand).

\begin{algorithm}
\caption{Pixel-Attentive Multi-Fingered Grasping}
\begin{algorithmic}[1]
  \STATE Initialize $zoom$ to $True$
  \STATE Initialize $\theta$ to trained model
  \STATE $s^{depth}, s^{scale} \leftarrow $ single depth map, 1
  \WHILE{zoom}{
    \STATE Sample action $a$ given $s^{depth}, s^{scale}$
    \STATE $zoom \leftarrow a^{zoom}$
    \IF {$zoom$}
        \STATE Crop original depth image around $a^{position}$ and at scale $(a^{scale} \times s^{scale})$ to acquire new $s^{depth}$
        \STATE $\text{New } s^{scale} \leftarrow s^{scale} \times a^{scale}$
    \ENDIF
    }
    \ENDWHILE
  \STATE Transform $a^{position}$ to Cartesian coordinates $\{x, y, z\}$ using point cloud inferred from depth map
  \STATE Move robot to joint positions defined by $\{x, y, z, a^{roll},$ $a^{pitch}, a^{yaw}, a^{spread}, a^{finger1}, a^{finger2}, a^{finger3}\}$ with a 5cm offset along target end effector orientation
  \STATE Close robot fingers at constant joint velocity until maximum effort and lift hand
\end{algorithmic}
\label{algo:algorithm}
\end{algorithm}
\subsection{Policy Optimization}
Let $\theta$ be the parameter weights of the entire network and $\pi_\theta$ be the RL policy the robot is trying to learn: $\pi_\theta: \mathcal{S} \to \Pi(\mathcal{A})$.
The robot's goal is to maximize the cumulative discounted sum of rewards: $\underset{\theta}{\text{maximize }}\mathbb{E}_{\pi_{\theta}}[\sum_{t}\gamma^{t-1} r_t]$.
The reward during the final timestep $t_{final}$ is a binary indicator of whether the robot successfully picked \textit{any} object up: $r_{t_{final}} = \mathds{1} \{\text{pick-up is successful}$\}. We follow the standard policy optimization objective:
\begin{equation}
  \underset{\theta}{\text{maximize }} \mathcal{L} = \mathbb{E}_{ \pi_{\theta}}[\pi_{\theta}(a_t \mid s_t) Q_{\pi_{\theta}}(s_t, a_t)]
\end{equation}

We opted out baseline subtraction for variance reduction since empirically it does not improve performance significantly. During zooming, no grasp is generated and $a_t$ is defined only by $\{a_t^{position}, a_t^{zoom}, a_t^{scale}\}$. During grasping, $a_t$ is defined by every component except $a_t^{scale}$. Therefore:
\begin{equation}
    \begin{split}
        &\log \pi_{\theta}(a_t \mid s_t) \\
        &= \log \pi_{\theta}(a_t^{position} \mid s_t) + \log \pi_{\theta}(a_t^{zoom} \mid s_t, a_t^{position})\\
        & + a_t^{zoom} \times \log \pi_{\theta}(a_t^{scale} \mid s_t, a_t^{position}) \\
        & + (1 - a_t^{zoom}) \times \sum_{dof \in DOFs}\log \pi_{\theta}(a^{dof}_t \mid s_t, a_t^{position}),
        \\&\text{where } DOFs = \{\text{roll}, \text{pitch}, \text{yaw}, \text{spread}, \text{finger1}, \\
        &\quad\qquad\qquad\qquad\quad \text{finger2}, \text{finger3}\}
    \end{split}\raisetag{2\baselineskip}
\end{equation}
In practice, we use the Clipped PPO objective~\cite{schulman2017proximal} to perform stable updates by limiting the step size\footnote{PPO Hyperparameters. Learning rate: $1 \times 10^{-4}$, number of epoches per batch: 10, number of actors: 14, batch size: 500, minibatch size: 96, discount rate ($\gamma$): 0.99, GAE parameter ($\lambda$): 0.95, PPO clipping coefficient ($\epsilon$): 0.2, gradient slipping: 20, entropy coefficient ($c2$): 0, optimizer: Adam}. Summarizing the above, the full Pixel-Attentive Multi-fingered Grasping procedure is shown in Algorithm~\ref{algo:algorithm}.

\subsection{The CNN Architecture}
Shown in Figure~\ref{algo}, the CNN architecture is inspired by Feature Pyramid Networks \cite{DBLP:journals/corr/LinDGHHB16}. During an episode, the input depth map is being ``zoomed in'' every timestep until the very last, therefore the CNN needs to have strong scale invariance (i.e. robustness against change in the scale of the scene objects), hence the feature pyramid blocks in the network.

\subsection{Rationale}
\subsubsection{Reinforcement-Learned Attention}
This algorithm design enables the robot to focus on a sub-region of the entire cluttered scene to grasp better \textit{locally}, or ``attention''. Without attention, the robot is presented with too much \textit{global} visual information in the cluttered scene such that it is difficult to grasp a \textit{local} object well. With attention, the robot can gradually zoom into the scene and focus on fewer and fewer objects as the episode continues. Since the task of generating good attention that will lead to good grasps \textit{in the future} and the task of generating a good grasp \textit{now} require similar reasoning around the objects' local geometry, \textit{one} single CNN branch $f^{position}$ can be trained to perform both tasks.

During training, the CNN receives upstream gradient signals encoding how successful the grasp was. Therefore, the CNN is trained to update its weights such that 1) it outputs a Position Map encoding a good grasp position if the episode terminates \textit{at the current timestep} and 2) it outputs the Attention Maps that zoom appropriately into the depth image to yield a good grasp when the episode terminates.
\subsubsection{Solving the Challenge of High Real-World Learning Sample Complexity}
While one can learn Pixel-Attentive Multi-Fingered grasping directly in real-world environments, this is inefficient without highly parallelized robot farms due to 1) the high sample complexity requirement of policy gradient methods and 2) the slow execution of real robots, and 3) the difficulty of generating near-i.i.d. cluttered grasping environments in the physical world. Instead, we opted to learn directly in simulation and transfer to the real environment \textit{without additional learning}. This high sim-to-real fidelity originates from the observation that the main sim-to-real gaps for vision-based learning come from texture (RGB) information, rather than depth information.

\section{Experiments} 
\begin{figure}[t]
\centering
    \includegraphics[width=\columnwidth]{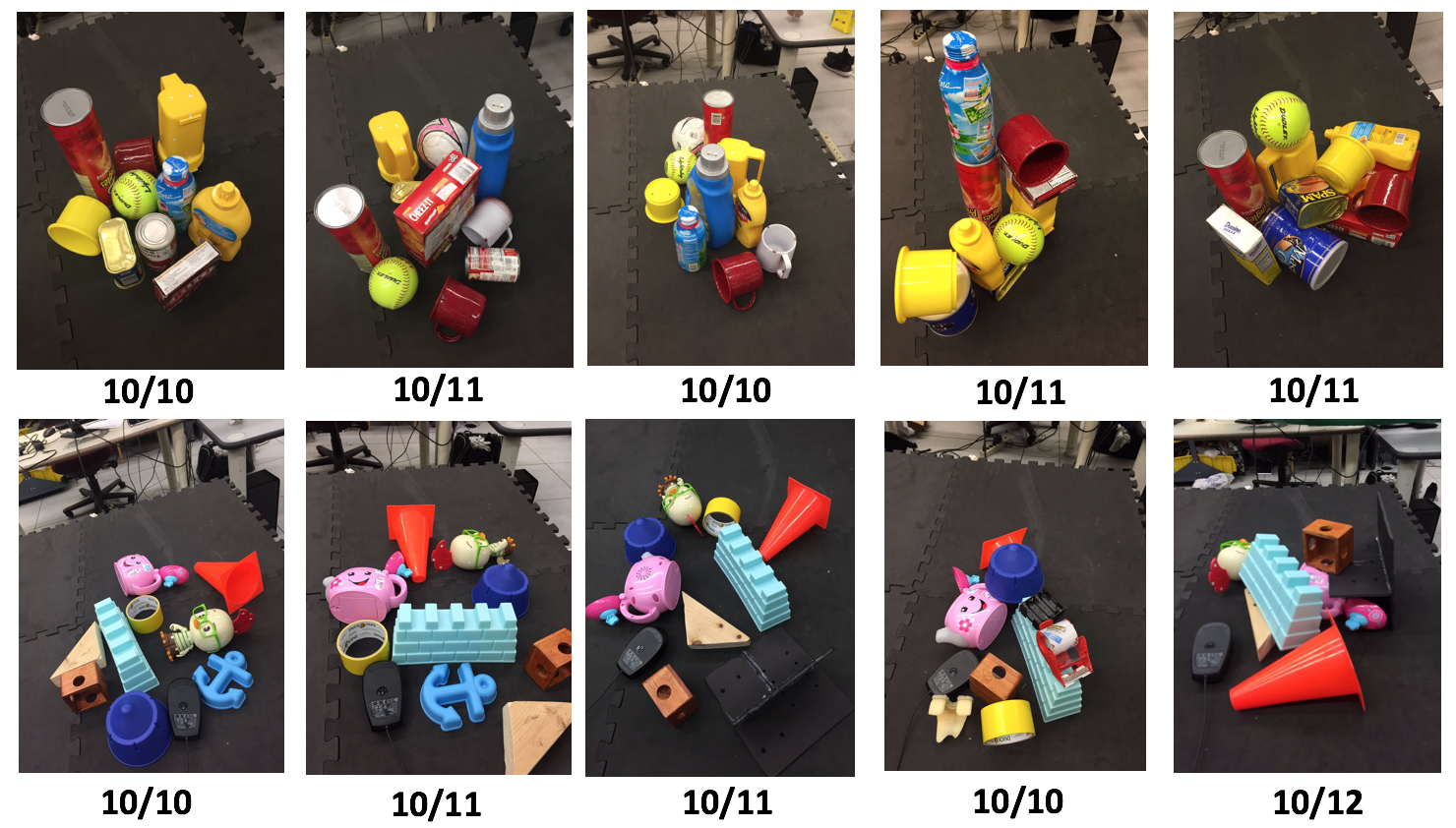}
\caption{\small Results for a few cluttered scenes used for our real-world experiments. In each scene, there are 10 objects randomly placed on the table. We report \# successful grasps / \# grasp attempts. For example, for the top right scene, the robot cleared the scene -- picking up all 10 objects with a total of 11 trials. \textbf{Top}: Scenes with seen objects similar to those used during training. \textbf{Bottom}: Scenes with novel objects different from those in training.}
\label{multi}
\vspace{-5mm}
\end{figure}

We train Pixel-Attentive Multi-Fingered grasping entirely in simulation and test in both simulation and real-world. During training, a single-object or multi-object cluttered scene is loaded with equal probability. We place one object in a single-object scene, a random number of objects from 2 to 30 for a simulated cluttered scene (Figure~\ref{fig:pb}), and 10 objects for a real-world cluttered scene. Using the ShapeNet Repository \cite{chang2015shapenet} in simulation, we use 200+ seen objects from the YCB and KIT datasets and 100+ novel objects from the BigBIRD dataset. We evaluate 500 grasp attempts per experiment in simulation. In real-world grasping, we use 15 YCB-like seen objects and 15 novel objects shown in Figure~\ref{multi}. We evaluate real-world single-object performance across 10 trials per seen/novel object, and real-world cluttered scene performance across 15 cluttered scenes. Video of the experiments can be found at \url{http://crlab.cs.columbia.edu/pixelattentivegrasping}.

\subsection{Results and Discussion}
\setlength\tabcolsep{5pt}
\begin{table}
\vspace{2mm}
\centering
\caption{Main Experiments and Ablation Results}
\label{tab:results}
\begin{tabular}{|c|cc|cc|}
\hline
 \multicolumn{1}{|c} {} & \multicolumn{2}{c|}{Single Object} & \multicolumn{2}{|c|} {Cluttered Scene}\\
\multicolumn{1}{|c} {Objects} & \multicolumn{1}{c} {Seen} & \multicolumn{1}{c|} {Novel} & \multicolumn{1}{c} {Seen} & \multicolumn{1}{c|} {Novel}\\ \hline
Ours (Sim) & 93.8 $\pm$ 2.6 & 94.9 $\pm$ 1.4 & 92.5 $\pm$ 1.8 & 91.1 $\pm$ 3.7 \\
Ours (Real) & 96.7 $\pm$ 6.2 & 93.3 $\pm$ 8.1 & 92.9 $\pm$ 5.8 & 91.9 $\pm$ 6.7\\ \hline
& \multicolumn{4}{|c|}{Ablation (Simulation)} \\ \hline
No Attention & 86.9 $\pm$ 4.7 & 85.2 $\pm$ 2.7 & 70.9 $\pm$ 6.3 & 72.2 $\pm$ 3.6\\
Top-Down & 88.6 $\pm$ 2.1 & 87.0 $\pm$ 2.7 & 74.8 $\pm$ 2.9 & 70.8 $\pm$ 5.9 \\
Parallel & 50.4 $\pm$ 6.7 & 44.5 $\pm$ 5.4 & 49.0 $\pm$ 2.4 & 45.1 $\pm$ 4.4\\
$60\degree$ Camera & 92.3 $\pm$ 2.1 & 91.9 $\pm$ 3.4 & 91.8 $\pm$ 3.6 & 91.6 $\pm$ 2.5\\\hline
\end{tabular}
\vspace{-3mm}
\end{table}
\subsubsection{Generalization to novel objects}
Shown in Table~\ref{tab:results} Row ``Ours (Sim)'', the test success rates in simulation for seen vs. novel objects are statistically similar for both single-object scenes ($93.8 \pm 2.6\%$ vs. $94.9 \pm 1.4\%$) and cluttered scenes ($92.5 \pm 1.8\%$ vs. $91.1 \pm 3.7\%$), exhibiting good transfer to novel objects. In Row ``Ours (Real)'', we notice similarly stable transfer performance to novel objects in the real world. 

The learned generalization to novel objects benefited from the partial observability of the MDP, discouraging the network from overfitting to seen objects. Since the depth map is the only input modality, the visual features are much less complex than that of complete 3D geometry, making the network select the safest grasp regardless of what ground-truth geometry it is beneath the point cloud.

\subsubsection{Generalization to real-world scenes}
Comparing Row ``Ours (Sim)'' against Row ``Ours (Real)'', we observe good real-world transfer given that no real-world training was done, mainly due to using depth as the only input modality, which has a smaller sim-to-real fidelity gap compared to texture/RGB information.
We show real-world performance of individual cluttered scenes of seen and novel objects in Figure~\ref{multi}. Note that the cluttered scenes include severe overlap and occlusion (two rightmost images of each row). 

\subsubsection{Cluttered scene performance}
Comparing Column ``Single Object'' to Column ``Cluttered Scene'', we observe good cluttered scene performance, which comes mainly from our attention mechanism and domain randomization, i.e. the random number of objects being placed into the scene during training. Under the attention mechanism, the network learns to focus on fewer and fewer objects as the episode continues. This eliminates perceptual distractions from objects in the rest of the scene that are far away from the object of interest. 

Qualitatively, most of the failure cases originates from objects being tightly cluttered with no spacing for the robot to insert its fingers. The robot ends up attempting to pick-up more than one object which results in the objects sliding out during lift. Such cases are very difficult to tackle unless the scene is perturbed. Since our method runs iteratively, the failure during the attempted lift produces necessary perturbation to the scene such that a successful grasp can be generated on the next try.

\subsection{Ablation}
\subsubsection{Importance of attention mechanism to performance}
We conducted experiments using a finite horizon of 1 instead of infinite horizon, effectively preventing the robot from using attention to zoom into the scene. Comparing Row ``No Attention'' to Row ``Ours (Sim)'', we observe larger performance degradation for cluttered ($21.6\%$ and $18.9\%$ for seen and novel objects respectively) than for single object scenes ($6.9\%$ and $9.7\%$ for seen and novel objects respectively) . Qualitatively, having no attention occasionally resulted in no learning. We attribute these findings mainly to the network's inability to pay attention to local regions of the cluttered scenes during training. 


\subsubsection{Using top-down grasp only}
By enforcing the value of $a_t^{pitch}$ to $\frac{\pi}{2}$, we restrict the robot to top-down only grasps. Comparing Row ``Top-Down'' against Row ``Ours (Sim)'' reveals larger performance degradation on cluttered scenes ($17.7\%$ and $20.3\%$ for seen and novel objects respectively) than for single-object scenes ($5.2\%$ and $7.9\%$ for seen and novel objects respectively), mainly because in cluttered scenes the robot needs non-top-down grasps to generate a better grasp on the target object that avoids nearby objects.

\subsubsection{Using parallel grasps only}
To examine the performance contribution of using multi-fingered grasps as opposed to two-fingered grasps, we enforce the lateral spread $a_t^{spread}$ to 0, effectively operating a two-fingered hand. Comparing Row ``Parallel'' to Row ``Ours (Sim)'' reveals performance degradation of $43.4\%$, $50.4\%$, $43.5\%$, $46.0\%$ on single-seen, single-novel, cluttered-seen, cluttered-novel scenes respectively, indicating a relatively significant contribution from using multi-fingered hands. Qualitatively, we observe frequent failures of grasping cylindrical or spherical objects. 

\subsubsection{Non-top-down camera viewing angle}
We tested our algorithm with a non-top-down camera viewing angle ($60\degree$). The simulation results in Row ``$60\degree$ Camera'' are statistically similar to Row ``Ours (Sim)'' (top-down view), showing robustness to different camera viewing angles.

\section{Conclusion} 
This work presents a novel way to reinforcement-learn high dimensional robotic grasping for multi-fingered hands without requiring any database of grasp examples. Using a policy gradient formulation and a learned attention mechanism, our method generates full 6-DOF grasp poses as well as all finger joint angles to pick-up objects given a single depth image. Entirely trained in simulation, our algorithm achieves $96.7\%$ (single seen object), $93.3\%$ (single novel object), $92.9\%$ (cluttered seen objects), $91.9\%$ (cluttered novel objects) pick-up success rate on the real robot and statistically similar performance in simulation, exhibiting good performance for real-world grasping, cluttered scenes and novel objects. In the future, we hope to apply our full 6-DOF grasping system to semantic grasping and combine object detection systems to achieve object specific grasping. 

\section*{ACKNOWLEDGMENT}
We thank Jake Varley and Wei Zhang for important advice. We also thank Jake Varley and David Watkins for developing the Barrett Hand controller, and everyone at Columbia Robotics Lab for useful comments and suggestions.

\bibliographystyle{IEEEtran}
\bibliography{IEEEabrv,references.bib}
\end{document}